\documentclass[acmsmall,screen]{acmart}
\usepackage[defblank]{paralist}
\usepackage{tikz,subfigure}
\usetikzlibrary{arrows,shapes,shapes.multipart,automata,backgrounds,petri,positioning,shadows,matrix,decorations.pathmorphing,fit,positioning,calc,backgrounds,arrows.meta}
\usepackage[section]{placeins}

\usepackage[final]{changes}

\definechangesauthor[color=lightgray]{del}
\definechangesauthor[color=gray]{add} 
\definechangesauthor[color=red]{delr2}
\definechangesauthor[color=blue]{addr2} 
\newif\ifshowcomments
\showcommentstrue

\ifshowcomments
\newcommand{\mynote}[2]{\fbox{\bfseries\sffamily\scriptsize{#1}}
 {\small$\blacktriangleright$\textsf{#2}$\blacktriangleleft$}}
\else
\newcommand{\mynote}[2]{}
\fi

\AtBeginDocument{%
  \providecommand\BibTeX{{%
    \normalfont B\kern-0.5em{\scshape i\kern-0.25em b}\kern-0.8em\TeX}}}

\setcopyright{acmcopyright}
\copyrightyear{2022}
\acmYear{2022}
\acmDOI{XXXXXXX.XXXXXXX}

\begin{document}

\title{\added[id=addr2]{AI-Augmented}\deleted[id=delr2]{Augmented} Business Process Management Systems: A Research Manifesto}


\author{Marlon Dumas}
\email{marlon.dumas@ut.ee}
\affiliation{%
  \institution{University of Tartu}
  \city{Tartu}
  \country{Estonia}
}
\email{marlon.dumas@apromore.com}
\affiliation{%
  \institution{Apromore}
  \city{Melbourne}
  \country{Australia}
}
\author{Fabiana Fournier}
\email{fabiana@il.ibm.com}
\author{Lior Limonad}
\email{liorli@il.ibm.com}
\affiliation{%
  \institution{IBM Research - Haifa}
  \city{Haifa}
  \country{Israel}
}
\author{Andrea Marrella}
\email{marrella@diag.uniroma1.it}
\affiliation{%
  \institution{Sapienza University of Rome}
  \city{Rome}
  \country{Italy}
}
\author{Marco Montali}
\email{montali@inf.unibz.it}
\affiliation{%
  \institution{Free University of Bozen-Bolzano}
  \city{Bolzano}
  \country{Italy}
}
\author{Jana-Rebecca Rehse}
\email{rehse@uni-mannheim.de}
\affiliation{%
  \institution{University of Mannheim}
  \city{Mannheim}
  \country{Germany}
}
\author{Rafael Accorsi}
\email{rafael.accorsi@ut.ee}
\affiliation{%
  \institution{University of Tartu}
  \city{Tartu}
  \country{Estonia}
}
\author{Diego Calvanese}
\email{calvanese@inf.unibz.it}
\affiliation{%
  \institution{Free University of Bozen-Bolzano}
  \city{Bolzano}
  \country{Italy}
}
\author{Giuseppe De Giacomo}
\email{degiacomo@diag.uniroma1.it}
\affiliation{%
  \institution{Sapienza University of Rome}
  \city{Rome}
  \country{Italy}
}
\author{Dirk Fahland}
\email{d.fahland@tue.nl}
\affiliation{%
  \institution{Eindhoven University of Technology}
  \city{Eindhoven}
  \country{The Netherlands}
}
\author{Avigdor Gal}
\email{avigal@ie.technion.ac.il}
\affiliation{%
  \institution{Technion}
  \city{Haifa}
  \country{Israel}
}
\author{Marcello La Rosa}
\email{marcello.larosa@unimelb.edu.au}
\affiliation{%
  \institution{The University of Melbourne}
  \city{Melbourne}
  \country{Australia}
}
\email{marcello.larosa@apromore.com}
\affiliation{%
  \institution{Apromore}
  \city{Melbourne}
  \country{Australia}
}
\author{Hagen V\"olzer}
\email{hvo@zurich.ibm.com}
\affiliation{%
  \institution{IBM Research - Zurich}
  \city{Zurich}
  \country{Switzerland}
}
\author{Ingo Weber}
\email{ingo.weber@tu-berlin.de}
\affiliation{%
  \institution{TU Berlin}
  \city{Berlin}
  \country{Germany}
}


\renewcommand{\shortauthors}{ABPMS Manifesto}

\begin{abstract}
\added[id=addr2]{AI-}Augmented Business Process Management Systems (ABPMSs) are an emerging class of process-aware information systems,  \deleted[id=delr2]{that draws upon} \added[id=addr2]{empowered by} trustworthy AI technology. An ABPMS enhances the execution of business processes with the aim of making these processes more adaptable, proactive, explainable, and context-sensitive. 
This manifesto presents a vision for ABPMSs and discusses research challenges that need to be surmounted to realize this vision.
To this end, we define the concept of ABPMS, we outline the lifecycle of processes within an ABPMS, we discuss core characteristics of an ABPMS, and we derive a set of challenges to realize systems with these characteristics.
\end{abstract}



\maketitle

\section{Vision and Motivation}
\label{sec:intro}

An increased availability of business process execution data, combined with advances in Artificial Intelligence (AI), have laid the ground for the emergence of information systems where the execution flows are not pre-determined, adaptations do not require explicit changes to software applications, and improvement opportunities are autonomously discovered, validated, and enabled on-the-fly. We call such systems \emph{\added[id=addr2]{AI-}Augmented Business Process Management Systems} (ABPMSs).

An ABPMS is a \deleted[id=delr2]{trustworthy and AI-empowered} process-aware information system that \added[id=addr2]{relies on trustworthy AI technology to} reason and act upon data, within a set of \deleted[id=delr2]{constraints and assumptions} \added[id=addr2]{restrictions}, \deleted[id=del]{derived from business process
knowledge and application domain knowledge,} with the aim to continuously adapt and improve a set of business processes with respect to one or more performance indicators. 


This definition takes as starting point the postulate that an ABPMS is a type of Information System (IS)
\cite{wand_weber_1995}, which maintains some representation, typically digital, of a domain (i.e., a part of the world). Following Alter \cite{alter_2008_is}, we view an IS as a ``work system whose processes and activities are devoted to [...] capturing, transmitting, storing, retrieving, manipulating, and displaying information''. 
\added[id=add]{Specifically, we consider an ABPMS as one with the following capabilities, 
which we illustrate below using a running example of a cold-chain transportation dispatch process for perishable goods.}
\deleted[id=del]{In addition to these basic elements of an IS, a \emph{process-aware IS} fulfills the following criteria:} 
\begin{enumerate}
\item It supports one or more processes\added[id=add]{, where a business process is \emph{``a collection of tasks that are executed in a specific sequence to achieve some business goal, such as producing a service or product for customers” ~\cite{Weske2012}}}. This property sets an ABPMS apart from systems that use AI to support isolated tasks. For example, an Optical Character Recognition (OCR) software can be used to produce a digital version of a paper invoice. This may be an important step in a process, but if the software only supports an isolated task, it is not process-aware.
\item It tracks the execution of the processes it supports, ensuring that each process conforms to a set of restrictions (e.g., policies)\added[id=addr2]{, herein called the \emph{process frame}.}\deleted[id=delr2]{and assumptions (e.g., anticipated process sequencing).} \deleted[id=delr2]{We use the term \emph{frame} to refer to the specification of such restrictions.} \added[id=addr2]{Following Minsky~\cite{minsky75}, we take as a starting point that a \emph{``frame is a data structure for representing a stereotyped situation, like being in a certain kind of living room [capturing] several kinds of information [...] about how to use the frame [...] what one can expect to happen next [and] about what to do if these expectations are not confirmed.''}. 
In the context of an ABPMS, these ``expectations'' are restrictions on possible states or actions, which the ABPMS is expected to abide to. In other words, the frame defines the boundaries within which the ABPMS should operate. These boundaries may be tightly or loosely specified. For example, a Petri net specifying the prescribed behavior of a process is an example of a tightly framed execution, while a set of temporal logic constraints that must hold true during the execution of the process constitutes a more loosely framed execution. Potentially, any approach presented in \cite{reichert2012} is a  candidate for framing the process in an ABPMS.} 
\item It orchestrates the activities of the processes it supports, such as, starting an activity once its preconditions are fulfilled\added[id=addr2]{, routing, or performing activities, within the given frame}. \deleted[id=del]{However, we do not presume that ABPMSs are based on explicit process specifications, e.g.\ graphical process models. Instead,} 
\added[id=addr2]{In other words, an ABPMS operates largely autonomously, within the boundaries set by the process frames (as it may manage multiple processes), while facilitating human-machine cooperation and guidance when the restrictions are not or cannot be met. The latter property implies that an ABPMS has a mechanism to detect and respond to situations where the restrictions in a process frame cannot be met. To this end,} an ABPMS may use AI technology, such as computer vision, signal processing, Natural Language Understanding (NLU), or knowledge graphs, to identify process-level changes to detect and respond to such situations. \deleted[id=del]{For example, in a cold-chain transportation dispatch system, a sensor in one of the trucks indicates a malfunctioning in its cooling. As a result, the system determines that there is enough time for a detour to the nearest service station and changes the planned route of the truck.}
\added[id=add]{In our example, a sensor in a truck indicates a cooling malfunction. The system then needs to find the nearest service station and reroute the truck there. It also needs to decide whether there is enough time to wait for the truck to be fixed or whether it needs to send a replacement truck to ensure the delivery of the goods in time. Each of these steps entails a change at the process level (insertion of activities).
}
\item
\added[id=add]{It uses AI technology to attain the goal of a business process within the given frame. Thus, a Business Process Management System (BPMS) that does not embed any AI capability is not an ABPMS.
On the other hand, a BPMS should not fully realize all the characteristics and lifecycle steps described in this paper to be an ABPMS, nor does it need to incorporate all the AI techniques mentioned in this paper. On the contrary, any BPM system that incorporates AI technology to drive the execution of business processes fits within the definition of an ABPMS.
}

\end{enumerate}


\deleted[id=del]{Besides being process-aware, an ABPMS is \emph{AI-empowered}. This means it employs AI technology (e.g., machine learning, NLU, automated reasoning). Thus, a Business Process Management System (BPMS) that does not embed AI capability is not an ABPMS.}

Embedding AI into a BPMS makes trust a crucial aspect of acceptance~\cite{Trust_AI_Y,europeancommission2020white}. In a traditional BPMS, each task or decision in a process is either driven by a human actor or by a software application that executes the task according to pre-determined business logic. In contrast,
in an ABPMS, the system automatically decides if a task is executed at all, and if so, how it will be executed, e.g.\ by a human agent or by the system, without pre-defined execution scripts. 
\deleted[id=del]{
Thus, \emph{trustworthiness} is crucial in an ABPMS. 
Mayer et al.~\cite{Trust_AI_67} define trust as: ``the willingness of a party to be vulnerable to the actions of another party based on the expectation that the other will perform a particular action important to the trustor, irrespective of the ability to monitor or control that other party''. 
Based on this definition, we postulate that for an ABPMS to be trustworthy, it should enable the users who control and monitor its operation to be convinced that they can rely on it. 
This requires the ABPMS to give faithful and tactically aligned \emph{explanations} and to allow its operators to ensure it is running all business processes as intended and preserving the ability to recover and retract from undesirable or inconsistent conditions.}


\added[id=add]{One may wonder whether an ABPMS is nothing but a general intelligence system that integrates one or more AI techniques (such as truth maintenance, knowledge graphs, machine learning, and automated reasoning). This is indeed the case, provided that the system monitors and orchestrates the execution of one or more business process(es). Thus, truth maintenance can be used to maintain and update knowledge about the process unfolding (in a broad sense); knowledge graphs can be exploited to store structural information about the objects and relations manipulated by the process; machine learning can be employed to predict and recommend what to do next, and so on and so forth.} 
\added[id=addr2]{However, the realization of an ABMPS requires the development and deployment of AI technologies that take into consideration the characteristics of (business) processes, chiefly:}
\begin{enumerate}
\item \added[id=addr2]{Their discrete temporal nature -- a business process consists of collections of activities, events and decisions that transform one or more objects to attain a certain goal.}
\item \added[id=addr2]{The imperative for compliance, i.e. staying within the frame}.
\item \added[id=addr2]{The ability to interact with business managers to pursue the goals of the process at the tactical level (e.g.\ optimizing the process performance indicators).}
\item \added[id=addr2]{The ability to coordinate the work of several stakeholders at an operational level, and to detect and respond to situations that potentially lead to situations outside the frame.}
\end{enumerate}

\added[id=add]{An orthogonal, key dimension is how much ``open-ended'' the ABPMS must be. This depends on the degree of predictability and repetitiveness of the process at hand, which may indeed affect which augmentation techniques can be applied, and how effective they can be.}

In this manifesto, we elaborate on the above definition of ABPMS by outlining a possible lifecycle of a business process that is executed, adapted, and continuously improved via AI technology. Based on this augmented lifecycle, we analyze the characteristics of an ABPMS, and contrast these characteristics to those of a conventional BPMS~\cite{PourmirzaPDG17}.


\section{AI-Augmented BPMS: A Lifecycle}
\label{sec:abpm_life_cycle}

\definecolor{marcofucsia1}{RGB}{203,41,123}
\definecolor{marcofucsia1}{RGB}{203,41,123}
\definecolor{marcofucsia2}{RGB}{153,25,94}
\definecolor{marcofucsia3}{RGB}{102,29,70}
\definecolor{marcoblue1}{RGB}{17,183,225}
\definecolor{marcoblue2}{RGB}{16,163,201}
\definecolor{marcoblue3}{RGB}{5,132,165}
\definecolor{marcoblue4}{RGB}{4,105,131}
\definecolor{marcoblue5}{RGB}{0,77,128}
\definecolor{marcoorange}{RGB}{255,147,0}
\definecolor{marcogreen1}{RGB}{82,174,139}
\definecolor{marcogreen2}{RGB}{56,122,101}

\definecolor{aicolor}{RGB}{0,77,128}

\tikzstyle{agent}=[
    fill=marcofucsia2!90,
    inner sep=0pt,
]

\tikzstyle{elem}=[
    font=\bfseries\sffamily,
]

\tikzstyle{lifecyclestep}=[
    elem,
    rectangle,
    rounded corners=8pt,
    minimum width=18mm,
    minimum height=10mm,
]

\tikzstyle{aistep}=[
  lifecyclestep,
  fill=aicolor!90,
  text=white,
]

\tikzstyle{bpmstep}=[
  lifecyclestep,
  fill=white,
  draw=marcofucsia2!90,
  densely dotted,
  line width=2pt
]

\tikzstyle{augstep}=[
  lifecyclestep,
  fill=white,
  draw=aicolor!90,
  line width=7pt
]

\tikzstyle{actor}=[
    rectangle,
    minimum width=11.5cm,
]

\tikzstyle{io}=[
  Straight Barb-Straight Barb,
]

\tikzstyle{seq}=[
  -Triangle,
  thick,
]

\tikzstyle{biseq}=[
  Triangle-Triangle,
  thick,
]

\pgfdeclarelayer{background}
\pgfdeclarelayer{foreground}
\pgfsetlayers{background,main,foreground}

\begin{figure}
\centering
\resizebox{\textwidth}{!}{
\begin{tikzpicture}[->,>=stealth',auto,x=15mm,y=1.0cm,node distance=20mm and 15mm,thick,font=\bfseries\sffamily]

 \node[
        agent,
        circle,
        minimum size=6mm,
    ] (head) at (0,0) {};

\node[
        agent,
        rectangle,
        rounded corners=5pt,
        below=.5mm of head,
        minimum width=10mm,
        minimum height=10mm,
    ] (body) {};

\node[ 
      agent,
      rectangle,
      rounded corners=4pt,
      below=.5mm of head,
      xshift=-1.5mm,
      minimum width=2.5mm,
      minimum height=20mm,
    ] (leg1) {};

\node[
      agent,
      rectangle,
      rounded corners=4pt,
      below=.5mm of head,
      xshift=1.5mm,
      minimum width=2.5mm,
      minimum height=20mm,
    ] (leg2) {};

\node[
    fill=white,
    rectangle,
    left=-\pgflinewidth of leg1.west,
    anchor=south east,
    minimum width=.5mm,
    minimum height=7.5mm,
    inner sep=0pt,
    yshift=-.5mm,
    ] (arm1) {};

\node[
    fill=white,
    rectangle, 
    right=-\pgflinewidth of leg2.east,
    anchor=south west,
    minimum width=.5mm,
    inner sep=0pt,
    minimum height=7.5mm,
    yshift=-.5mm,
    ] (arm2) {};

  \node[
    below=10mm of body.south,
    anchor=north,
    font=\bfseries\sffamily\large
  ] (agentlabel) {AGENT};

  \node[
    below=2.45cm of body.south,
    xshift=-5mm,
    bpmstep,
    minimum width=7mm,
    minimum height=7mm,
    rounded corners=4pt,
  ] (bpmlegend) {};
  
  \node[
    right=0mm of bpmlegend
  ] (bpmlegendlabel) {traditional BPM};

  \node[
    above=1mm of bpmlegend.north west,
    anchor=south west,
    xshift=-2mm
  ] {task types};
  
  \node[
    below=1mm of bpmlegend.south,
    aistep,
    minimum width=9mm,
    minimum height=7mm,
    rounded corners=4pt,
  ] (ailegend) {};
  
  \node[
    right=0mm of ailegend
  ] (ailegendlabel) {AI};

  \node[
    below=1mm of ailegend,
    augstep,
    minimum width=6.5mm,
    minimum height=5mm,
    rounded corners=2pt,
  ] (auglegend) {};
  
  \node[
    right=0mm of auglegend
  ] (auglegendlabel) {AI-augmented};

 \begin{pgfonlayer}{foreground}

  \node[
    augstep,
    right=7.2cm of body
  ] (frame) {frame};
  
  \node[
    bpmstep,
    minimum width=15mm,
    minimum height=7mm,
    rounded corners=2pt,
  ] at (frame.center){frame};
  
  \node[
    augstep,
    below=1cm of frame,
    xshift=-12.5mm,
  ] (enact) {enact};
  
  \node[
    bpmstep,
    minimum width=15mm,
    minimum height=7mm,
    rounded corners=2pt,
  ] at (enact.center){enact};
  
  \node[
    augstep,
    right=5mm of enact,
  ] (perceive) {perceive};
  
  \node[
    bpmstep,
    minimum width=15mm,
    minimum height=7mm,
    rounded corners=2pt,
  ] at (perceive.center){perceive};
  
  \node[
    augstep,
    below=5mm of perceive,
    xshift=-1.25cm,
  ] (reason) {reason};
  
  \node[
    bpmstep,
    minimum width=15mm,
    minimum height=7mm,
    rounded corners=2pt,
  ] at (reason.center){reason};
  
   \draw[seq] (perceive) edge (reason);
   \draw[seq] (reason) edge (enact);
   \draw[seq] (enact) edge (perceive);
  
  \node[elem,below=1mm of reason] (execute) {process-aware execution};
  
\end{pgfonlayer}

 \begin{pgfonlayer}{main}    
  \node[
    fit=(enact) (perceive) (reason) (execute),
    lifecyclestep,
    fill=marcoorange!50,
    rounded corners=14pt,
    draw=darkgray!90,
    ultra thick,
    densely dashed,
  ] (cycle) {};
  
  \node[
    aistep,
    right=10mm of cycle,
  ] (explain) {explain};
  
  \node[
    aistep,
    left=5mm of cycle,
    yshift=7.5mm,
  ] (adapt) {adapt};

  \node[
    aistep,
    left=5mm of cycle,
    yshift=-7.5mm,
    xshift=-10mm,
  ] (improve) {improve};

  \draw[biseq] (frame) edge (cycle);
  \draw[biseq] (explain) edge (cycle);
  \draw[biseq] (adapt) edge (cycle);
  \draw[biseq] (improve) edge (cycle);
  \draw[seq,rounded corners=10pt] (explain) |- (frame);
  \draw[seq,rounded corners=5pt] (adapt) |- ($(frame.west)-(0,3mm)$);
  \draw[seq,rounded corners=10pt] (improve) |- (frame.west);


  
  

\end{pgfonlayer}

\begin{pgfonlayer}{background} 
  \node[
    fit=(cycle) (improve) (frame) (explain),
    fill=marcoorange!20,
    draw=gray!90,
    ultra thick,
    densely dotted,
  ] (abpms) {};
  
  \node[
    left=1mm of abpms.south east,
    yshift=1.5mm,
    anchor=south east, 
    font=\bfseries\sffamily\large
  ] {ABPMS};
  
  \draw[io,aicolor,line width=4pt] (.25,-1.5) -- (1.95,-1.5);
  \draw[io,white,line width=2pt,densely dashed] (.3,-1.5) -- (1.9,-1.5);
  
  \node[
    elem
  ] at (1.1,-0.8) {\begin{tabular}{c}intelligent\\interaction\end{tabular}};

  \node[
    fit=(ailegend) (auglegend) (ailegendlabel) (auglegendlabel) (bpmlegend) (bpmlegendlabel),
    draw=black,
    thick
  ] {};

\end{pgfonlayer}

\end{tikzpicture}
}
\caption{The ABPMS lifecycle. Rounded rectangles represent lifecycle steps, while arrows denote sequencing.}
\label{fig:lifecycle} 
\end{figure}
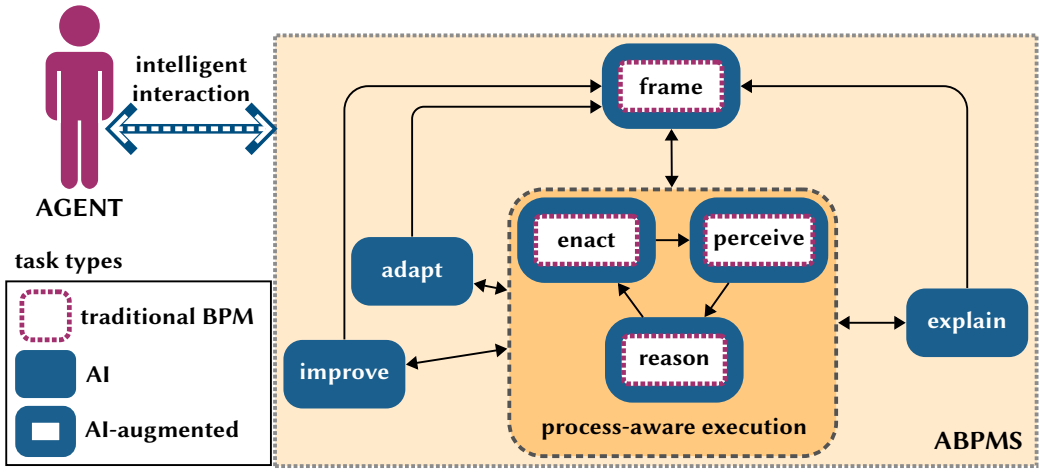

In this section, we focus on the operational lifecycle of an ABPMS, as shown in Figure~\ref{fig:lifecycle}. We describe the different steps, which in turn require the ABPMS to exhibit certain characteristics, discussed in detail in Section~\ref{sec:background}. We distinguish between two step types: 
\begin{itemize}
    \item \emph{Basic steps} (frame, enact, perceive, reason). These are steps that can be found also in the lifecycle of a traditional BPMS \cite{DLMR18}, but that are here augmented with AI technologies.
    \item \emph{Advanced steps} (explain, adapt, improve) that are specific to ABPMSs, and only exist when AI is an integral part of the system.
\end{itemize}

The lifecycle involves two main actors: the ABPMS itself, and one or more \emph{agents} that interact with the ABPMS to achieve their \emph{goals} while complying with \emph{process constraints}. In this context, the term ``agent'' has to be interpreted as an umbrella term for any autonomous entity entitled to use and control the ABPMS, whether human or digital. 
\deleted[id=del]{
The ABPMS \emph{autonomously} executes its lifecycle, while \emph{continuously interacting} with its agents. The latter implies that the ABPMS must have  \emph{conversational} capabilities. 
}

The lifecycle starts when an agent \emph{frames} the ABPMS, that is, equips the ABPMS with initial constraints and goals. Constraints may incorporate different types of knowledge, such as predefined procedures, norms, regulatory constraints, commonsense rules, and best practices. The frame defines the boundaries within which the ABPMS should operate. How the frame is specified (e.g., whether it is interpreted and verified symbolically, or used to rewards or other indirect compliance mechanisms) depends on the AI techniques that the ABPMS incorporates. In any case, since an ABPMS seeks to optimize a set of measurable \added[id=add]{business process goals}, these constraints are likely to refer to Key Performance Indicators (KPIs). Such KPIs enable the ABPMS to determine whether the goals are met, or to estimate how distant it currently is from a goal state. Note that an ABPMS may have multiple competing goals and may be tasked with finding an optimal trade-off between them. 

In terms of business process specification languages, framing can employ a wide repertoire of different modeling languages, including:
\begin{inparaenum}[\it (i)]
\item traditional graphical process modeling languages employing such as BPMN \cite{DLMR18}.
\item flexible process modeling languages \cite{reichert2012}, including
declarative approaches such as Declare \cite{PeSA07,MPAC10} and DCR graphs \cite{MuHS13} equipped with temporal logic-based semantics \cite{MukH10,DeDM14};
\item ``hybrid'' combinations of these languages (see, e.g., \cite{SaSO01,SSMR16,DDVP16}).
\end{inparaenum}

\added[id=addr2]{We acknowledge that the concept of `framing', as presented by Minsky~\cite{minsky75}, is traditionally conceived as an underlying data stereotyping mechanism or a set of facts about particular objects and event types and arranging the types into a large taxonomic hierarchy or a knowledge graph. In an ABPMS, we build on the core idea of framing as the set of articulations that accounts for process awareness in the system and sets the maximal permissive boundary for its execution. As a general concept, a frame does not have to conform to a single linguistic or symbolic formalism in its realization and may cater a combination of element specifications such as goals to be achieved, transition rules (e.g., legal and illegal states and transitions), constraints and prohibited execution scenarios, different strategies and rewards for execution, and performance metrics. Overall, a frame is a core characteristic in ABPMSs, defined as the holistic set of epistemic articulations that ensures the system inherently retains (some degree of) process awareness in its execution cycle (namely, `process aware' execution as illustrated in Figure~\ref{fig:lifecycle}). As such, the frame lets the ABPMS determine how it can progress with its executions, whether it is just set to discover a process, to monitor a given process, to alert about process mis-conformances, or to (autonomously) drive the full operations of say a certain business unit. We purposely keep the concrete specification of a frame open, to give the freedom of its realization to employ any viable means of articulation, ranging anything between a rigid process model (i.e., traditional BPMS) on the one hand to a general intelligence AI that is trained to maintain process awareness on the other.}

The actual process executed by the ABPMS is in general not directly obtained from the provided constraints, but instead emerges from the ABPMS itself in later stages of the lifecycle. At the same time, different modes of execution for the ABPMS can be instrumented depending on how the system is framed.
At one extreme of framing we have the case where an agent instructs the ABPMS with a very specific goal (\deleted[id=del]{such as maximize profit and minimize time-to-delivery in an order-to-cash scenario} \added[id=add]{such as to minimize time-to-delivery in our running example, neglecting cost and potential disturbances to the cold chain}), and a detailed set of strict constraints yielding an imperative process (such as a BPMN definition of the \deleted[id=del]{order-to-cash} \added[id=add]{dispatch} process). \added[id=add]{In this latter scenario, where an imperative modeling notation is used}, framing is essentially equivalent to the modeling step in the traditional BPMS lifecycle. At the other extreme of framing, we have the situation where only the goals are given, leaving the ABPMS completely free to decide how to best achieve them. 
\added[id=add]{The goal structure embedded in the ABPMS -- externally through framing and internally through learning -- will then determine its interactions with and implications on the world around it, particularly with regard to the resources it utilizes.
If the freight of the cold chain truck is so valuable that the ABPMS is not allowed to take any risks, it will have the cooling repaired immediately, even though it might increase the time-to-delivery. If the ABPMS’s goal is to minimize time-to-delivery, it might notice that the truck is close enough to its destination and the ambient temperature is still acceptable, so it continues the delivery and repairs the truck afterwards.
}

Once an initial framing is completed, the lifecycle enters its central stage, namely \emph{process-aware execution}. This consists of the rotation between perceiving, reasoning, and enactment. Such rotation serves as the intrinsic mechanism that accommodates for process-awareness, continuously tracking the \added[id=add]{business process} state, and ensuring its execution respects the corresponding framing. 

During this stage, the system \emph{autonomously} executes a \emph{process} that emerges from the ABPMS itself depending on (continuously updated) goals and constraints, as well as additional extensional and intentional information obtained by the ABPMS during execution. Internally, this stage is structured following the well-known \emph{sense-think-act} cycle, a widely adopted paradigm to define architectures of dynamic AI systems \cite{russell_2016_ai}. 

Process-aware execution is substantiated by three main steps. First, the ABPMS \emph{perceives} the environment and acquires new data concerning the process and the current state of affairs. Such data may be heterogeneous, with different levels of granularity and degrees of certainty. For example, the ABPMS may acquire event data directly from a workflow management system, from structured legacy systems \cite{CKMS17}, and/or from unstructured textual and visual sources \cite{KnPS20}. Such data may be incomplete and uncertain. For example, when using a video camera to detect relevant event data, it may not be straightforward to distinguish whether a clerk is inspecting a package or loading it into a truck, and the recorded view may be partial. 

Although the perceived data often refer to the extensional level, i.e., to facts that the ABPMS perceives to be true about the world, they may also pertain to the intentional level, i.e., the process enacted by the ABPMS. Examples of extensional data of the first kind are the creation of a new order by the customer or the detection of a defect in a package. Examples of intentional data of the second kind are a new inspection procedure for packages that must be acquired by the ABPMS, or the refinement of a previously acquired constraint.

Transforming collected data into information, extracting relevant events, pondering about their uncertainty, and combining them with the process, constraints and goals towards deciding which actions to take next is the purpose of the second step, namely \emph{reasoning}. This pertains to a wide repertoire of relevant AI techniques that may be employed by the ABPMS, such as data integration, event/state recognition, knowledge (graph) construction, planning, simulation, reasoning about actions and goals, verification, estimation, and prediction of the short and long-term effects and rewards of actions.

Next, the actual execution is carried out in the consequent, \emph{enactment} step, where the ABPMS uses its own actuators to interact with the environment. This interaction may occur directly or indirectly.
For example, the ABPMS may autonomously create and send an invoice to a customer, if it has direct access to the ERP system of the organization; if it has no access, it may alternatively converse with a human operator from the sales department or with an external software agent to perform the action. The latter scenario typically characterizes physical actions to be directly performed in the world, in case the ABPMS is not equipped with physical actuators.  

Once an enactment step is performed, a new perception step is typically carried out, sensing the induced effects of the action and starting a new iteration of the cycle.

At any step during the execution cycle, the ABPMS system may decide to perform one of the advanced steps: provide \emph{explanations} about the past, current, and anticipated states of the system, \emph{adapt} itself to new circumstances and drifts, as well as \emph{improve} (e.g., optimize) its execution against its goals, the available resources, and the framing constraints. The execution of those tasks could lead the ABPMS to revise and update its internal knowledge both at the extensional level, by adding and revising information about objects, relations, facts and events, and at the intentional one, by evolving its internal model of the process, as well as its constraints and goals. In other words, the ABPMS may autonomously \emph{reframe} itself on the basis of the newly acquired knowledge and a corresponding adaptation, improvement, or explanation step, while respecting the previous frame (cf.~Section~\ref{sec:framed-autonomy}). Importantly, reframing can also be carried out by the agents themselves, depending on the current circumstances and the continuous evolution of goals and requirements. 

\added[id=addr2]{Implied from the notion of framing, we can thus distinguish an ABPMS from any other form of an AI-empowered system where its core intelligence mechanism is realized without having an explicit component of (process-aware) framing as a key characteristic of an ABPMS. As intelligent and accurate such a system may be, and whether or not it exploits any of the recent advances in AI technologies (e.g., NLP, DL, RL), we consider such an AI realization to be an ABPMS only if it possesses at least the process execution steps running under the framing to govern for process awareness.}


\section{Characteristics of an ABPMS}
\label{sec:background}

To realize the ABPMS lifecycle, we postulate that an ABPMS should be:  
\begin{inparaenum}[(1)]
\item \emph{(framed) autonomous} to act independently and proactively;
\item \emph{conversationally actionable} to seamlessly interact with agents whenever necessary;
\item \emph{adaptive} to react to changes in its environment;
\item \emph{(self-)improving} to ensure the optimal achievement of its goals;
\item \emph{explainable} to ensure the trust and hence the cooperation of the human agents. 
\end{inparaenum}

\subsection{(Framed) Autonomous}
\label{sec:framed-autonomy}
One characteristic that is inherent, yet implicit, in the presented lifecycle is that an ABPMS acts \emph{autonomously} within the provided frame. This means that the system performs the lifecycle steps independently and proactively. ABPMSs can leverage AI capabilities to assess the circumstances of a specific situation and then independently select and execute the most appropriate action. This is particularly apparent for the reasoning step, where ABPMSs can draw on either symbolic or sub-symbolic AI capabilities to find its own patterns and make deductions.

The degree of autonomy depends on the framing, which determines the system's actions in two ways. 
First, framing acts as a maximal permissive boundary for the execution of the ABPMS. It provides the goals and constraints within which the process is executed. As stated in Section~\ref{sec:abpm_life_cycle}, the ABPMS may independently reframe itself if it acquires new knowledge during adaption, improvement, or explanation. However, the designer may restrict the ABPMS's autonomy to reframe itself. We refer to this capability as \emph{meta-framing}. Meta-framing gives designers the option to determine which parts of the frame are modifiable and which can only be modified via human instructions. 
\deleted[id=del]{For example, an ABPMS may be allowed to autonomously change the price by up to 10\% but has to ask permission for any larger changes.}
\added[id=add]{For example, an ABPMS may be allowed to accept deviations in temperature by up to 10\% but has to notify a human agent in case of larger changes.}
Meta-framing allows designers to control the degree of autonomy of the ABPMS, from a system that acts fully autonomously to one that is never allowed to reframe itself. This facilitates a gradual buildup of trust on the part of the human agents.

Seeing the process enacted by the ABPMS as a maximally permissive behavior confined within the boundaries of the current frame is reminiscent of supervisory control \cite{CaLa08,WonhamCai2019} and synthesis \cite{EhlersLTV17} in discrete-event systems. Our setting is arguably more complex due to the nature of the frame, which includes both hard constraints, soft constraints, and different types of goals, in addition to a continuous evolution and adaptation of the frame itself.

The second implication of the framing on the degree of autonomy relates to the abilities of the ABPMS to capture its environment. The potential actions that the ABPMS may take depend on the information that is provided during framing. The more information the ABPMS has about the context of the process and its underlying constraints and assumptions, the more it is able to make its own informed decisions and act accordingly. 
Ideally, the ABPMS will be \emph{adjustably autonomous}, i.e., able to recognize areas where it needs more information, and then ask one of its agents to provide it. 

\subsection{Conversationally Actionable}

An ABPMS is expected to act largely autonomously, reducing the involvement of human agents. However, those agents are still an essential part of the ABPMS lifecycle. Hence, being able to interact with humans effectively is a central characteristic of an ABPMS \cite{muthusamy_2020_dos}. Depending on the level of autonomy, the ABPMS may ask for input or guidance to frame the process appropriately. In return, the human agent may ask the ABPMS for some  facts. The agent might also provide the ABPMS with new insights about the execution context. These situations require direct interaction between the ABPMS and a human agent. 

To keep the human effort at a minimum, we postulate that an ABPMS should be \emph{conversationally actionable}, i.e., able to proactively communicate with human agents about process-related actions, goals, and intentions, using natural language possibly enhanced with richer interfaces. This characteristic ensures that the ABPMS seamlessly integrates into the workday of the human agents. 
Here, ``conversational'' means that the ABPMS offers interfaces for both written and spoken natural language to cater to the users' preferences and to facilitate both synchronous and asynchronous communication. In addition, it can adapt its mode of conversation to both the preferences of the user, the type of information that needs to be communicated, and the situation in which the communication happens.

``Actionable'' means that the ABPMS makes concrete recommendations to the user and engages in a discussion about their benefits and drawbacks. For example, if a manager enters the office in the morning, they could verbally ask the ABPMS for a recap of the previous day's process performance. The ABPMS could then present the requested information as an interactive dashboard and recommend inspection of some erroneous cases. The manager can interact with the dashboard, ask the ABPMS for potential options to mitigate those problems, and engage in a discussion to eventually find counter-measures. 

\subsection{Adaptive}

Business processes are often embedded in dynamic contexts characterized by uncertainty and change~\cite{baiyere2020digital}. ABPMSs must be able to operate under these circumstances. 
For example, a person who was competent to perform an activity yesterday may no longer be certified for its execution today, such that the ABPMS has to reassign the activity without disrupting the process execution. A
\deleted[id=del]{machine that performed some manufacturing activity}
\added[id=add]{cooling truck}
may have broken down, such that the ABPMS has to find a substitute, or the truck has become temporarily unavailable such that the APBMS has to redirect or slow down process flows to avoid cascading interruptions~\cite{ToosinezhadFKA20}. The company may acquire new knowledge, such as an expert's opinion, or new tools, such as AI technology, that enable the execution of previously unknown process steps. 

An ABPMS must therefore be able to \emph{adapt} itself to the new process context in real time. It needs to notice changes in its environment and react appropriately. In contrast to conventional BPMSs, where ad-hoc changes to individual process instances may be performed only via human interventions~\cite{dadam2009adept}, an ABPMS may trigger adaptations autonomously. 
\deleted[id=del]{Thus, the ABPMS should be able to perform sequences of ad-hoc inferencing activities to resolve unforeseen issues. 
These inferences may be hard to rationalize in retrospect without explicit considerations of time, and of its unfolding via a series of beliefs, justifications, and without faithful state tracking.} 

\subsection{Self-Improving}

Whereas the adaptive nature of an ABPMS refers to its capabilities to react to external changes, an ABPMS should also proactively anticipate changes and be constantly \emph{self-improving}. This means that an ABPMS should predict changes and be optimally prepared for those changes. 
\deleted[id=del]{For example, an ABPMS can detect the gradually decreasing precision of a production machine and predict that it will have to undergo maintenance operations within the next weeks. It can then schedule those operations for a day when the predicted workload is low and reserve capacities on other machines to take over the concerned process step. }
\added[id=add]{For example, an ABPMS can detect the gradually increasing temperature in a cooling truck and predict that it will have to undergo maintenance operations within the next weeks. It can then schedule those operations for a day when the predicted transportation load is low and reserve capacities in other trucks to take over the concerned orders.}
\deleted[id=del]{If an organization wants to make a decision about, e.g., the pricing of a product, the ABPMS can feed data into a machine learning model trained on historical data on the implications of previous pricing decisions and suggest a price in real-time.}

The objective of the ABPMS is to strike optimal trade-offs between the competing performance measures of its process, such as operational cost and waiting times~\cite{lopez-pintado2021optimization}. Therefore, it autonomously decides what needs to be done to achieve such trade-offs. These decisions may affect the whole process or individual cases and be temporary or permanent.
\deleted[id=del]{In the above example of a production process, the ABPMS can either schedule the maintenance immediately or wait for a day with a low production workload, weighing a longer production time and higher cost against a higher output quality.}
\added[id=add]{For our example, the ABPMS can either schedule the maintenance immediately or wait for a day with a low transportation load, weighing a longer delivery time and higher cost against a higher output quality.}
\deleted[id=del]{Finding optimal trade-offs under consideration of the many influencing factors on the future performance of a process can only be achieved through powerful AI-based prediction and simulation technology \cite{dumas2021constructing}.}


\subsection{Explainable}

\added[id=add]{
Acceptance of an ABPMS requires users to trust that the system will execute the process in a way that they find rational and justified.
Thus, \emph{trustworthiness} is crucial characteristic of an ABPMS. 
Mayer et al.~\cite{Trust_AI_67} define trust as: ``the willingness of a party to be vulnerable to the actions of another party based on the expectation that the other will perform a particular action important to the trustor, irrespective of the ability to monitor or control that other party''. 
Based on this definition, we postulate that for an ABPMS to be trustworthy, it should enable users to understand its choices. 
This requires the ABPMS to give faithful and tactically aligned \emph{explanations} and to allow its operators to ensure it is running all business processes as intended and preserving the ability to recover and retract from undesirable or inconsistent conditions.}

\deleted[id=del]{To this end, Explainability for AI (XAI) aims to highlight decision-relevant features in the algorithmic model that either contribute to the model accuracy or to a specific prediction for one particular observation \cite{rehse2019xai}. This includes \emph{post-hoc} methods for feature-dependence quantification such as backpropagation and perturbation-based techniques \cite{post-hoc-XAI} or model agnostics techniques such as SHAP \cite{shap}.}

\deleted[id=del]{While explainability is often agnostic to the particular algorithmic model, the model itself may possess a topology that helps depicting the rationale for its results. This property is typically referred to as \emph{model interpretability} (\emph{ante-hoc} methods), reflecting the level at which a given model makes sense for a human observer [...] .}
\deleted[id=del]{Explainability of the reasoning step can be realised by a variety of models (ante- post-hoc). 
Furthermore, it may stem from the intrinsic property of the specific AI models and from the global property of the business process model if known. 
However, the current state of the art is only local to the context in which AI is employed along the overall (global) reasoning process \cite{vinod1} and thus, is lacking process-awareness when employed in ABPMSs.}

\deleted[id=del]{Explainability is one of several properties required to achieve trustworthiness.}
Toreini et al. \cite{Trust_AI_X} identify four aspects of trustworthiness: Fairness, Explainability, Auditability, and Safety (FEAS). Fairness technologies focus on detection or prevention of discrimination and bias in different demographics; explainability technologies focus on explaining and interpreting the outcome to the stakeholders (including end-users) in a humane manner; auditability technologies focus on enabling third parties and regulators to supervise, challenge or monitor the operation of the model(s); safety technologies focus on ensuring the operation of the model as intended in presence of active or passive malicious attacker.
In the scope of an ABPMS, we view explainability in its business broader sense, also governing for fairness and other regulatory concerns (hence, auditability), whereas the safety aspect is addressed in the framing step.

In light of the above, we consider process-aware explainability as an inherent property of an ABPMS. An ABPMS should be equipped with the ability to independently and continuously reason about process enactment outcomes (a 'second-tier' of reasoning), most often retrospectively. This includes ongoing capturing of key conditions (e.g., historical framing that reflect timely assumptions and beliefs) and the ability to draw inferential associations (dependencies) between such conditions and intermediary process execution results (i.e., reasoning and/or enactment). Such drawing helps to autonomously establish/quantify the situational validity of any derived process output. Constantly tracking such inferential associations could be used to discover invalid or insignificant results and may also be a basis for the ability to retract process executions. 

\deleted[id=del]{XABPMSs apply XAI techniques, taking into account the BP sequencing semantics as a means to streamline series of inferential reasoning associations.}

\section{Research Challenges and Opportunities}
\label{sec:challenges}

Enabling ABPMSs requires overcoming a handful of research challenges. Some of the more prominently identified research directions are the following:

\subsection{Situation-aware Explainability} 

The vision about ABPMSs, as opposed to BPMSs, is that such a system possesses intrinsic explainability capabilities (cf.\ the `explain' step in fig. \ref{fig:lifecycle})\deleted[id=del]{that is intended to act upon situations that cannot be `explained'}. `Explaining' is used here as a general term, which could unfold itself into a variety of articulated queries\added[id=add]{, such as: ``what are the reasons for performing task T? why was decision X taken? when was it concluded that the business process goal had been attained?''} As an inherent step, explainability should be realised to have its output presented to the agent, so to have the agent is able to understand and accordingly act upon it and the evolving situation inducing it (e.g., retract to a point in time when the inconsistency could be eliminated). Existing techniques lack in many cases clear semantics of their reasoning and remain detached from broader process context. For example: consider a patient already awaiting admission to a surgery while a new lab result arrives with some new metrics that reverse the need to operate. Will the system be able to actively stop the process and retract to the point at which surgery may need to be re-assessed? 

Providing explanations on why a machine state is reached has been recently investigated, motivated by the opaqueness of so called “black‐box” approaches \cite{Holzinger2018,Holzinger2019}. Finding an appropriate explanation is not easy, because this requires understanding the context and providing a description of causality and consequences of a given fact. Frequently, explanations cannot be derived from ``local" inference (e.g., current undergoing task or decision in a business process), but require explicit bookkeeping that links the current execution frame to its corresponding chain of causal inferential associations, allowing traceability and trackability to the (root) causes that contributed to its deduction. Maintaining this type of historical inference associations is necessary to provide explanations at the process ("global") level. Jan et al. ~\cite{vinod1} motivate the need for process-aware explanations. Here, process-aware explainability refers to taking advantage of the knowledge of the business process definitions and full runtime process traces for better explanations. 

The embedding of AI in the context of BP enables considering richer, time-related, contextual information that relates to application of each AI capability during some process execution. While time might be the most prominent factor affecting a system's state, other factors including location (spatial) and objects grouping (claim process behaves differently for different types of customers), and combinations of these, are also of importance and should be taken into consideration when analyzing different outcomes and behaviors.

Our proposed research direction goes beyond taking into account the sequencing imposed by the process, and includes a much broader reasoning context in the lifecycle of an ABPMS which is responsible for capturing the relevant situational factors beyond the mere local context. Hence, our desire in an ABPMS is to enable a broad, \emph{situation-aware explainability}, one that is able to capture traces of reasoning highlights (i.e., combination of `reasoning' parts and corresponding framing), and is able to reflect potential inferential associations that go beyond the local reasoning context. Further to this, situation-aware explainability also entails ongoing tracking of inferential associations between subsequent enactments, as a basis to gaining confidence in its ability to retract upon inconsistencies, and handle a large variety of situations.

As with the emergence of Reason Maintenance Systems (e.g., \cite{doyle1979truth}) in relation to knowledge bases back in the 70s, the consideration of AI in the context of processes presents a unique opportunity for further developments of new explainability techniques and process execution management infrastructure. Complemented by contemporary advances in causal reasoning (e.g., \cite{why-Peral,causal-Pearl}), such infrastructure could be developed to determine plausible justifications for process decisions and results (intermediary or eventual) in real-time, to allow for valid establishment of reasoning in retrospect. Specifically, \emph{event knowledge graphs}~\cite{EsserF21} which encode behavioral and causal inter-dependencies of objects and actors over time in the context of process flows and process knowledge allow to symbolically represent situations of all kinds for \emph{situation-aware reasoning}. Such techniques may be used to facilitate the (automatic or by humans) tracking of execution consistency, for better understanding of process flows and process outcomes, and to drive ongoing process improvements (at either design- or retraction at run-time).

\subsection{Augmented Process Automation}

An ABPMS should be able to achieve more complex automation than a traditional BPMS by using AI to minimize human-dependent training and supporting human users to execute complex tasks that entail decision making. A partial solution to this issue is provided by Robotic Process Automation (RPA) technology \cite{van2018robotic}. RPA is intended to operate on the user interface (UI) of software applications by creating software robots that automate mouse and keyboard interactions to perform repetitive tasks previously done by humans on the UI. This minimizes human error due to mental lapses resulting from boredom or exhaustion. A critical component to the usage of RPA is related to the identification of opportunities for automation to add the RPA technology in the right place and maximize its potential \cite{leno2021robotic}. 

Even though the research in RPA shows promising methods for assessing automation opportunities \cite{bosco2019discovering,Agostinelli2020,leno2020automated}, in practice tasks amenable to be automated are identified manually by subject matter experts. In addition, identifying automatic tasks only solves part of the problem. In fact, many tasks do not reflect routine work, and different users may adopt varying work practices for the same task, making it difficult to realize full process automation, which is not always feasible from a technical point of view.

Differently from RPA, an ABPMS does not aim at fully replacing the human users working on a process, but rather at leveraging AI techniques to augment the humans' capabilities and stimulate their tacit knowledge for executing process tasks in a trustworthy manner. A way to address this challenge is to rely on the concept of \emph{Hybrid Process Intelligence} \cite{van2021hybrid}, which postulates that an AI component must interact with human users as a ``learning apprentice'', adapting itself to the users' work practices (and not vice versa). To realize this vision, a two-sided interaction is required between the ABPMS and the human user. On the one hand, an ABPMS must be capable of recognizing previously unnoticed situations, and seamlessly escalate key decisions about individual cases or groups of cases to human decision makers, providing them with contextualized information and scenario assessments to support effective decision-making. On the other hand, human users' can always override a decision taken by the ABPMS to prevent it from making mistakes. Notably, the corrective actions performed by the human should be leveraged by the ABPMS to enhance its internal learning of the process and improve its autonomy.





\subsection{Automated Process Adaptation}

%
%

The management of processes in dynamic contexts requires that an ABPMS provides real-time monitoring and automated adaptation features during process execution to adapt processes to exceptions, exogenous events and any contextual change that may happen at run-time, by preserving their structure and minimising any human intervention. 

Initial research work addressing the need of adaptive process management can be traced back to the late 90s and early 2000s~\cite{eder1995,eder1996,casati1999,hagen2000,klein2000,luo2000,casati2001}. Abstracting from the specific techniques and implementations, a common behavioral pattern can be identified. At design-time, a process designer identifies possible exceptions that may occur and specifies suitable exception handlers to catch and fix the exceptions. At run-time, errors, constraint violations and other events might interrupt the process flow. The exception is detected and thrown, and an exception handler is invoked to catch the exception. \deleted[id=del]{Not surprisingly, this pattern resembles exception handling mechanisms in programming languages.}

\added[id=add]{A similar principle has been applied in YAWL \cite{ter2009modern}, a workflow language supported by a traditional BPMS implementation, where for each exception that can be anticipated, it is possible to define an exception handling process, named \emph{exlet}, which includes a number of exception handling primitives (for removing, suspending, continuing, completing, failing, and restarting a workitem/case) and one or more compensatory processes in the form of \emph{worklets} (i.e., self-contained YAWL specifications executed as a replacement for a work item or as compensatory processes). Exlets are linked to specifications by defining specific rules in the shape of RippleDown Rules specified as ``if condition, then conclusion,'' where the condition defines the exception triggering condition, and the conclusion defines the exlet. Notably, the BPMS supporting YAWL can be potentially extended to incorporate more advanced adaptation logic towards addressing the requirements of an ABPMS}.

In an attempt to increase the level of user support, the ADEPT system and its evolutions~\cite{adeptflex,adept,AristaFlow} rely on the possibility of performing structural ad-hoc changes over process instances at run-time. Structural adaptation techniques have been systematized through the identification of adaptation patterns~\cite{weber2008}, i.e., predefined change operations for adding, deleting or replacing process activities. Similarly, semi-automated approaches using case-based reasoning have been proposed in \cite{Weber@ACBR2004}. They exploit available knowledge about previously performed changes, so that users can retrieve and apply it to adapt a process. 

While in the previous work the degree of automation is generally limited to manual ad-hoc changes performed by experienced users, an ABPMS is in need of techniques that go beyond hard-coded solutions that put all the burden on IT professionals, which often lack the needed knowledge to model all possible contingencies at the outset, or this knowledge can become obsolete as process instances are executed and evolve, by making useless their initial effort. 

In this direction, a number of techniques from the field of AI have been applied to BPMSs with the aim of increasing the degree of automated process adaptation \cite{10.1007/978-3-319-74030-0_1,marrella2019automated}.
In~\cite{Weske@ADBIS2004,Ferreira@IJCIS2006,marrella2013synthesizing,DBLP:journals/soca/MarrellaL17}, if a task failure occurs at run-time and leads to a process goal violation, a new complete process definition that complies with the goal is generated relying on a partial-order AI planner. As a side effect, this often significantly modifies the assignment of tasks to process participants.
The work~\cite{Bucchiarone@SOCA2011} proposes a goal-driven approach to adapt processes to run-time context changes. Process and context changes that prevent goal achievement are specified at design-time and recovery strategies are built at run-time through an adaptation mechanism based on service composition via AI planning. 
Recently, the SmartPM approach \cite{MarrellaTIST2017,DBLP:journals/aicom/MarrellaMS18} has been proposed. It relies on a planning-based mechanism that requires no predefined handler to build on-the-fly the recovery procedure required to adapt a running process instance. Specifically, adaptation in SmartPM is seen as reducing the gap between the expected reality, i.e., the (idealized) model of reality that reflects the intended outcome of the task execution, and the physical reality, i.e., the real world with the actual values of conditions and outcomes. If a recovery procedure is needed during process execution (this happens if the two realities are different from each other), SmartPM invokes an AI planner to build a recovery procedure as a plan, which can thereby resolve exceptions that were not designed into the original process.

SmartPM, which can be considered as an implementation of the digital twin paradigm \cite{SEMERARO2021103469}, provides an important demonstration of how automated adaption can be incorporated into an ABPMS. However, SmartPM requires the design-time definition of the family of tasks involved in a process, annotated with pre-and post-conditions expressed in terms of data objects and attributes, and when changes to them become relevant. This may specifically require situation-aware descriptions of tasks in the context of complete, partially specified, under-specified process fragments of desired or forbidden behavior over the history and expected future evolution of all involved data objects and actors~\cite{FahlandW10,FahlandP12,HeweltW16}. Conversely, in an ABPMS, extensions are needed to incorporate new data types and tasks during run-time and to permit richer kinds of data and knowledge in the planning, which can be a daunting task. To easy the achievement of this task, a possible solution is to employ specific design patterns and heuristics at design-time \cite{SaSO01,Marron2020} that may help to improve the system's ability to deal with unanticipated situations at run-time.

\subsection{Perspective Agility}


\deleted[id=del]{Conventional BPMSs require us to predefine a model of the process which prescribes its detailed execution flow at run-time. On the other hand, an ABPMS is required to support process execution, where the process structure is unknown (or only partially known) at design-time and may emerge directly from the ABPMS itself at run-time.}


In contrast to BPMSs where the process model is predefined, ABPMSs should support processes with unknown (or incomplete) structure at design-time~\cite{FahlandW10,FahlandP12,HeweltW16} that may emerge at run-time. To realize this vision, an ABPMS must support a mixture of formalisms (e.g., imperative, declarative, goal-oriented, and actor routines and habits) and interrelated artifact types (e.g., temporal constraints, goal specifications, flowcharts, and data objects) and behaviours to achieve agile process executions. 

The challenge of multi-perspective support of processes thereby has to be addressed in two aspects: (1) integrating various formalisms, conceptualizations, symbolic, and sub-symbolic representations of processes and execution flows and (2) integrating behavioral characteristics and constraints of entities interacting in the shared process context. 

The first aspect was tackled in the BPM literature mainly in the modelling \cite{la2011configurable}, monitoring \cite{jalali2013multi} and mining \cite{GUZZO2021114934} phases, even if always targeting the integration of imperative and declarative formalisms only. The concept of perspective agility as a run-time issue is (partly) addressed by the philosophy behind Case Management and Modeling Notation (CMMN) standard \cite{wiemuth2017application}. CMMN provides the possibility for ad-hoc sequencing that emerges by the case.  Another alternative to CMMN could be to rely on the belief-desire-intention (BDI) agent perspective~\cite{georgeff1998belief}. In BDI, an agent is described by its beliefs (i.e., the information an agent has about itself and
its environment), its desires (i.e., motivations of the agent that drive its course of action), and its derived intentions (i.e., the short-term plans that the agent wants to execute). In an ABPMS, a process can be executed considering the current goal and the context to determine the next step of the process, and the agent can be seen as an assistant of the user who is responsible for driving a task through the process, whose real structure is discovered only during the process enactment. CMMN and BDI are only two examples. We envision that drawing explicit linkages among formalisms (e.g., through ontology mappings) can help synchronize the specification of different formalisms in a process-oriented way, towards multi-perspective support during process execution.

The second aspect has initially been researched to study the interplay of processes and data objects for automated reasoning. To this end, data-centric process models have been investigated, enriching persistent data models with process-aware dynamics \cite{CaDM13,DHLV18}. In addition, declarative and procedural process modeling formalisms have been extended with data inspection and manipulation capabilities. This led to a flourishing series of approaches, ranging from case variables and decisions \cite{BaHW17,FeLM21,LeFM21} to process networks that co-evolve multiple objects involved in one-to-many and many-to-many relationships \cite{Fahland19,AKMA19,PWOB19,GGMR20,FaDA21,DBLP:journals/is/0001PBFW15}, which called for novel process mining techniques \cite{Aals19, DBLP:journals/tsc/LuNWF15}. Recently, also the behavioral influence of actors and resources on the processes they perform has been investigated from the angle of performance. This research led to integrating queueing models and process models~\cite{SenderovichWYGM16} and the detection of complex performance patterns~\cite{DenisovFA18,KlijnF19}. This also allowed to increase accuracy in process prediction~\cite{SenderovichBGW19,KlijnF20}, inferring otherwise unobservable behavior~\cite{FaDA21} and detecting emergent system-level phenomena~\cite{ToosinezhadFKA20}.
Further, integrating explicit behavioral descriptions of process executions and actors allows to detect complex task execution patterns describing organizational routines and individual habits that involve multiple actors and process executions that evolve over time~\cite{KlijnMF21}.
All these approaches have in common that they integrate two or more ``behavioral dimensions'' or explicitly consider a multitude of observations in one dimension, e.g., all cases passing through a queue. What emerges is the need to explicitly distinguish different kinds of objects as well as their ``trajectories'' over time. We envision that a simple, but flexibly extensible graph-based model (e.g.,~\cite{EsserF21}) are capable of encoding the concepts and phenomena so far studied separately in a uniform format. 

Any solution to perspective agility thereby fundamentally has to offer the capability to describe and allow inference of the process in any chosen (combination of) perspectives.

\subsection{Actionable Conversations}

\deleted[id=del]{Reducing the need for direct human involvement in a business process is one of the key features envisioned by an ABPMS. To achieve that, we state that conversational interfaces with business processes may represent an interesting solution to facilitate the interaction between an ABPMS and the users, providing them with a natural language tool with a low learning curve to interact with.}
\added[id=add]{A major challenge to realize the vision behind an ABPMS consists in developing solutions (possibly with a low learning curve) that facilitate the interaction between an ABPMS and the users.}
Nowadays, there is a strong industry trend towards automating processes using reactive conversational agents (e.g., chatbots)  \cite{lopez2019process,moiseeva2020multipurpose}, which rely on simple scripts that drive users through a series of predefined questions. Conversely, an ABPMS should provide an interface, which relies on AI to create dynamic conversations that not only responds to user queries and performs actions on their behalf, but also initiates conversations with users in order to inform them of the process progression, alert them of relevant process changes (e.g., changes in demand distribution, customer behavior, or resource performance and availability), and make recommendations for interventions in order to improve performance with respect to relevant performance targets.

To develop such proactive actionable conversations with an ABPMS, we posit that an integrated usage of Natural Language Processing and Machine Learning techniques should be adopted, utilising their ability to infer conversation meaning from relevant vocabulary and remember previous conversations with users, enabling tailored responses for recurrent users. In this direction, some recent papers touched on the importance of natural language understanding in business process automation. Among them, in \cite{galitsky2019developing}, the authors present an approach towards less reactive chatbot development that relies on natural language understanding and generative machine learning models, emphasizing the importance of ``synthesis from examples''. While a significant overhead and domain knowledge are necessary to implement this approach, its predictable and controllable behavior makes it a robust candidate to be adopted for enterprise chatbots. In \cite{rizk2020unified}, the authors present a multi-agent framework to develop a conversational assistant supporting the capability of autonomously executing tasks in a business process, although many relevant challenges still need to be tackled, such as scalability, agent overlap (i.e., as the number of agents in the framework increases, some agent functionality and knowledge may overlap) and access control (some agents must not be accessible to specific users).


%
%

\section{Conclusion}
\label{sec:epilogue}

The introduction of AI technology into BPMSs creates a range of opportunities to exploit automation in business processes to make them more resourceful, with minimal, yet effective, engagement with human agents during their execution. These opportunities require a significant shift in the way the BPMS operates and interacts with its operators (both human and digital agents).
While traditional BPMSs encode pre-defined flows and rules, an ABPMS is able to reason about the current state of the process (or across several processes) to determine a course of action that improves the performance of the process.
To fully exploit this capability, the ABPMS needs a degree of autonomy. Naturally, this autonomy needs to be framed by operational assumptions, goals, and environmental constraints.
Also, ABPMSs need to engage conversationally with human agents, they need to explain their actions, and they need to recommend adaptations or improvements in the way the process is performed.

This manifesto outlined a number of research challenges that need to be overcome to realize systems that exhibit these characteristics. We cannot bring a proper closure to this manifesto as it stands in contrast with its purpose of being a trigger for future research. The postulated characteristics of an ABPMS\added[id=add]{, hence also the challenges,} are not exhaustive \added[id=add]{and others may be added}. \added[id=add]{They enable a modular progression from BPMS to ABPMS prioritised by the designer's choice.} They should rather be grasped as a first step towards shaping the notion of ABPMSs. Similarly, the stated challenges should be seen as an open call to design, develop, and validate methods and techniques that contribute to achieve the vision of an ABPMS as a system that exploits the capabilities of AI technology to support the continuous improvement of business processes during and as part of their execution.

\begin{acks}
Work supported by the European Research Council via Advanced Grants PIX (834141) and WhiteMech (834228).
\end{acks}

\bibliographystyle{ACM-Reference-Format}
\bibliography{acm-main}

\end{document}